\documentclass{article}
\usepackage[utf8]{inputenc}

\usepackage{url,hyperref,lineno,microtype,subcaption}
\usepackage[onehalfspacing]{setspace}

\def\Authors{Frida Heskebeck\,$^{1,*}$, Carolina Bergeling\,$^{2}$ and Bo Bernhardsson\,$^{1}$}

\usepackage{enumerate}
\usepackage{booktabs}
\usepackage{amsmath}
\usepackage{amssymb}
\usepackage{xcolor}
\usepackage[nameinlink,noabbrev,capitalise]{cleveref}
\DeclareUnicodeCharacter{2212}{-}

\hypersetup{
	allcolors=blue!40!black,
	colorlinks=true,
	pdftitle={MABsInBCIs},
	pdfauthor={Frida Heskebeck},
}
\DeclareMathOperator*{\argmax}{arg\,max}

\title{Multi-Armed Bandits in\\Brain-Computer Interfaces} 

\author{\Authors} 

\usepackage{authblk}

\author[1,*]{Frida Heskebeck}
\author[2]{Carolina Bergeling}
\author[1]{Bo Bernhardsson}
\affil[1]{Department of Automatic Control, Lund University, Lund, Sweden}
\affil[2]{Department of Mathematics and Natural Sciences, Blekinge Tekniska Högskola, Karlskrona, Sweden}
\affil[*]{frida.heskebeck@control.lth.se}
\date{}

\begin{document}
\sloppy

\maketitle
Paper under review, year 2022.
\begin{abstract}
The multi-armed bandit (MAB) problem models a decision-maker that optimizes its actions based on current and acquired new knowledge to maximize its reward. 
This type of online decision is prominent in many procedures of Brain-Computer Interfaces (BCIs) and MAB has previously been used to investigate, e.g., what mental commands to use to optimize BCI performance. 
However, MAB optimization in the context of BCI is still relatively unexplored, even though it has the potential to improve BCI performance during both calibration and real-time implementation. 
Therefore, this review aims to further introduce MABs to the BCI community.
The review includes a background on MAB problems and standard solution methods, and interpretations related to BCI systems. 
Moreover, it includes state-of-the-art concepts of MAB in BCI and suggestions for future research. 

\tiny
\section*{Keywords:} Multi-Armed Bandits, Brain-Computer Interfaces, Reinforcement Learning, Calibration, Real-Time Optimization 
\end{abstract}

\section{Introduction}

The multi-armed bandit (MAB) problem, introduced by Robbins in 1952 \cite{robbins_aspects_1952}, models an agent (decision-maker) that wishes to optimize its actions with the goal of maximizing the expected reward from these actions. 
The agent must decide between multiple competing actions based on only partial knowledge of their expected rewards and only gains new knowledge after an action is taken.
In other words, the agent has to \emph{explore} the action-space before it has enough knowledge to start to \emph{exploit} the learned best action. 
This procedure is known as the exploration vs. exploitation trade-off, which is nowadays often recognized from reinforcement learning. 
The MAB problem is, in fact, one of the simplest forms of reinforcement learning \cite{sutton_multi-armed_2018} and has been applied to many different fields of research, such as healthcare, finance, and recommender systems \cite{bouneffouf_survey_2019}. 
The name `multi-armed bandit' comes from a casino setting, where the agent, at each round, chooses one out of a given number of slot machines to play, with the objective to maximize the total payoff over time \cite{sutton_multi-armed_2018}. 

The aim of this paper is to review the MAB framework in the context of Brain-Computer Interfaces (BCIs). 
The exploration vs. exploitation tradeoff exists naturally within the procedures of BCI systems, such as deciding which data or paradigm to utilize for a particular task. 
It is especially so in the online setting, where properties of different choices might only be partially known but become better understood as more data is gathered. 

It is assumed that the reader is familiar with the BCI-field and we refer to, e.g., \cite{nam_braincomputer_2018-ch1} or \cite{nicolas-alonso_brain_2012} for any of the BCI-related nomenclature used in the paper.  
In  \cref{sec:MAB}, MABs are introduced as well as the algorithms often used to solve them. 
\cref{sec:BCI+MAB_today} highlights existing examples of MABs in the context of BCIs, while 
\cref{sec:BCI+MAB_future} provides suggestions for future research. 
Finally, some MAB programming packages are listed in \cref{sec:coding} and the paper is concluded in \cref{sec:conclusion}.

\section{Multi-Armed~Bandits theory - a crash course}
\label{sec:MAB}

\subsection{The MAB problem formulation}

The MAB problem is described as: at each time instant $t$, an agent chooses an action $a_t$ out of $K$ possible actions and receives a reward $r_{a_t}$. 
In a BCI setting, MABs could be used to optimize calibration data collection for motor imagery (MI) experiments as in \cite{fruitet_bandit_2012}. 
Then, $t$ corresponds to the next time for data collection, $K$ to the available MI classes, $a_t$ to the class for the next data collection, and $r_{a_t}$ to the increase of classification accuracy when retraining the classifier with the newly gathered data.
The reward for each action is not known beforehand. Moreover, the rewards are not deterministic, or fixed, but governed by some probability distribution. This means that the agent needs to perform an action $a$, often multiple times, in order to gain enough knowledge to accurately estimate or predict the reward $r_{a}$ \cite{sutton_multi-armed_2018}.  

The aim in a MAB problem is to design a strategy, or policy, for the agent on how to choose the actions such that the total reward after $T$ time steps, i.e., $\sum_{t=1}^T r_{a_t}$, is maximized. 
The policy is based on the agent's gathered knowledge from previous actions. 
The time horizon $T$, also called the agent's budget, is always finite in practice. However, when theoretically analyzing MAB problems, an infinite time-horizon, $T \rightarrow \infty$, is often assumed \cite{burtini_survey_2015}. 

In the original MAB problem the rewards are stationary with a binary distribution; 1 or 0, win or lose, with a probability $\theta_a$ of a win \cite{robbins_aspects_1952}. 
A beta distribution (see, e.g, \cite{faisal_probability_2020}) is often used to describe the distribution of $\theta_a$ (different actions have different beta distributions). \cite{scott_modern_2010}. 
An estimate of the probability to win with an action, $\hat{\theta}_a$, can for instance be sampled as $\frac{\alpha_a}{\alpha_a + \beta_a}$ where $\alpha_a$ and $\beta_a$ are the number of wins and losses for that action, respectively. The certainty of the estimate increases with the number of samples.  

Another common assumption on the rewards' distribution is Gaussianity, see \cite{faisal_probability_2020} for a definition. 
The reward can then take any value, not only 0 or 1. 
Each action has an unknown true mean $\mu$ for the reward and a standard deviation $\sigma$. 
Upon receiving a reward, the agent can update the estimated values $\hat{\mu}$ and $\hat{\sigma}$ \cite{sutton_multi-armed_2018}.

The MAB problem can be varied in multiple ways. For instance, the probability distributions of the rewards $r_{a_t}$ can be considered to be stationary or  changing over time. The set of possible actions $K$ can be fixed or non-fixed. The reward distributions could change depending on contextual information, and the policy of the agent needs not be restricted to one action at a time. 
In \cref{tab:MAB_cat} we illustrate some of the most common variants of MAB problems: the so-called original MAB problem, restless and switching bandits, mortal and sleeping bandits, contextual bandits as well as dueling bandits. 

\begin{table}[htbp]
\caption{Overview of MAB variants and their characteristics compared to the original MAB problem.}
\label{tab:MAB_cat}
\begin{tabular}{p{0.43\linewidth}  p{0.57\linewidth}}
\toprule
\textbf{Multi-Armed Bandit variant} & \textbf{Characteristic}                                  \\ \midrule
Original MAB problem                & Static reward and fixed set of actions                   \\
Restless and Switching bandits      & Non-static reward                                        \\
Mortal and Sleeping bandits         & Set of available actions changes                         \\
Contextual bandits                  & Rewards change based on state of surrounding environment \\
Dueling bandits                     & Agent chooses two actions at each time step              \\ \bottomrule
\end{tabular}
\end{table}

\subsection{Algorithms for solving MAB problems}
\label{sec:MAB_alg}

The aim for all algorithms, also called policies, is to balance the exploration vs. exploitation of the actions \cite{sutton_multi-armed_2018}.
Here, we present the most common algorithms, in the context of the original MAB problem formulation. 
There are two commonly used criteria when evaluating algorithms:

\begin{enumerate}[i) ]
\item
The \emph{cumulative reward}, also called the gain, $G_\phi(T)$, is the total reward over the time horizon $T$, averaged over multiple trials, i.e. \cref{eq:cumulative_reward}, where $r_{a_t}$ is the reward at each time step using the policy $\phi$ \cite{mahajan_multi-armed_2008}.

\begin{equation}
\label{eq:cumulative_reward}
    G_\phi(T) = \mathbb{E} \left[ \sum_{t=1}^{T} r_{a_t} \right]
\end{equation}

\item
The \emph{regret}, $R_\phi(T)$, is the difference between the total reward for the best action and the agent's total received reward over the time horizon $T$. In \cref{eq:regret}, $r^*$ is the best achievable reward, i.e., the expected reward for the best action, and $r_{a_t}$ is the agent's received reward at each time step using the policy $\phi$. 
The theoretical (upper) bounds on the regret, meaning the worst-case expected regret after $n$ number of plays, are often compared for different policies. 
If the regret bound is logarithmic, the optimal action is found with the policy. 
Analysis of the lower bounds on the regret shows the best case for finding the optimal action \cite{burtini_survey_2015}.

\begin{equation}
\label{eq:regret}
    R_\phi(T) =  T r^* - \mathbb{E} \left[ \sum_{t=1}^{T} r_{a_t} \right]
\end{equation}

\end{enumerate}

\subsubsection{Random policy}
In the random policy, the agent takes a random action at each time instance. 
This policy is often used as a baseline when comparing policies -- a policy should not be worse than the random policy. 

\subsubsection{$\epsilon$-greedy policy}
The agent gets an initial estimate of each action's reward by performing each action once. In a greedy policy, the agent always chooses the action with the highest estimated reward.
This method only exploits and never explores after the initial phase. 
If the agent's initial reward estimates are off, the policy will be stuck in always choosing a non-optimal action, 
giving a linear regret growth.

In the $\epsilon$-greedy policy on the other hand, the agent chooses the best action but with an $\epsilon$ probability picks a random action, \cite{sutton_multi-armed_2018}. 
The occasional random action forces the agent to explore all actions, which helps the agent to better estimate the actions' rewards so the agent can exploit the best action. 
Using the $\epsilon$-greedy policy, the agent can eventually guess quite well what action is best, but will still take a random action with $\epsilon$ probability. 
In that case, the random action will force the agent to act non-optimally, which is unwanted. 
Gradually decreasing $\epsilon$ over time reduces the probability of such non-optimal actions. Theoretically, such $\epsilon$-decreasing policies can be constructed which guarantee logarithmic  bounds on regret  \cite{burtini_survey_2015}, which is a significant improvement over linear growth.

Another variant of the $\epsilon$-greedy policy is the $\epsilon$-first policy. 
It requires the time horizon $T$ (how many actions the agent will be allowed to take in total) to be known beforehand. 
The agent takes a random action for the first $\epsilon T$ time steps and picks the action with the highest estimated reward for the remaining $(1-\epsilon) T$ steps. 
This policy has proven to be superior to the $\epsilon$-greedy policy when the time horizon is known and the rewards are stationary \cite{burtini_survey_2015}. 

\subsubsection{Upper Confidence Bound (UCB)}
In the Upper Confidence Bound (UCB) algorithm, the agent looks at the estimated reward plus an extra margin based on the uncertainty of the reward's estimate. 
The extra margin is calculated from the number of actions that have been taken in total and the number of times that action has been taken.
The algorithm for the next action $a_t$ is mathematically described as \cref{eq:UCB} where $\hat{r}_a$ is the estimated reward for that action, $t$ is the current time step, $n_a$ is the number of times the action has been taken and $c>0$ is a parameter \cite{sutton_multi-armed_2018}.

\begin{equation}
    \label{eq:UCB}
    a_t = \argmax_a \left[ \hat{r}_a + c \sqrt{\frac{\ln{t}}{n_a}}\right]
\end{equation}

The UCB algorithm does not have any assumption on the distribution of the rewards, and its regret is logarithmically bounded, as proven in \cite{auer_finite-time_2002}. 
There are many variants of the UCB algorithm that cope with non-stationary rewards or contextual information, such as LinUCB, Adapt-Eve, DiscountedUCB, and SlidingWindowUCB \cite{burtini_survey_2015}. 

\subsubsection{Thompson sampling}
Thompson sampling, also called probability matching, is an algorithm for MABs with binary rewards. 
The idea is to match the probability of choosing an action to its probability of being the best action. 
Mathematically this is quite advanced, but in practice, it means that the agent samples an estimated reward, $\hat{\theta_a}$, from each action's beta distribution and chooses the action with the highest such $\hat{\theta_a}$. 
The theoretical regret bound is logarithmic \cite{burtini_survey_2015}. 

\section{Current use of Multi-Armed~Bandits in Brain-Computer~Interfaces}
\label{sec:BCI+MAB_today}

To our knowledge, there is a limited use of multi-armed bandits in BCI systems today. 
We have found two main applications, described in the following subsections. 

\subsection{One button BCI -- Improving Calibration} 
\label{sec:one_button_BCI}
\cite{fruitet_bandit_2012} have a BCI system with one button that the user can press by Motor Imagery (MI) movements, e.g., imagining moving the right hand \cite{pfurtscheller_dynamics_2010}.
Different motor imagery tasks are optimal for different users and might also differ between sessions. 
Fruitet et al. aim to improve the calibration of such systems by focusing data collection on MI tasks with high performance rather than collecting data for all MI tasks, as in uniform calibration.
In their MAB problem formulation, the set of actions $K$ correspond to the available MI tasks, the time-horizon $T$ to the total number of data samples to collect, the action $a_t$ to the MI task of the following data sample to collect, and the reward $r_{a_t}$ to the classification rate of the corresponding MI task. 
The goal for MAB problems is to maximize the total reward, while the goal for Fruitet et al. is to maximize the classification rate of the optimal MI task.
Despite the slight goal difference, the exploration vs. exploitation trade-off is the same, and Fruitet et al. have based their algorithm on the UCB algorithm. 
They report higher classification rates with their algorithm than the uniform calibration approach. 
In a follow-up paper, \cite{fruitet_automatic_2013}, they try their algorithm in an online setting and proves it to be more efficient than the uniform calibration approach, confirming their findings in the first paper.

\subsection{Multi-Armed~Bandits in P300 spellers -- Real-time Implementations}

In a P300 speller, the letters are arranged in a grid, and a P300 signal is elicited when the row/column with the target letter is highlighted \cite{rezeika_braincomputer_2018, riggins_p300_2020}. 
In the paper \cite{ma_adaptive_2021}, they use Thompson sampling to shorten the time for finding the target letter by reducing the number of non-target row/column highlights.
In their MAB problem formulation, the set of actions $K$ correspond to the available stimuli groups of letters to highlight, the action $a_t$ to the next group, and the reward $r_{a_t}$ (being 0 or 1) to whether the selected group contained the target letter or not. 
The reward for each action follows a beta distribution where $\hat{\theta}_a$ represents the probability of the action's corresponding stimuli group containing the target letter. 
Their algorithm selects and evaluates multiple actions in each iteration, in contrast to classical MAB algorithms that select one action at each step.
They use a pre-defined stopping criterion rather than a fixed time-horizon $T$. 
They conclude that the use of MABs improve the performance of the BCI system.

There are multiple variants of MABs in P300 spellers, e.g., \cite{kocanaogullari_optimal_2018} and \cite{guo_calibration-free_2021}. 
The MAB problem formulation in \cite{kocanaogullari_optimal_2018} is similar to \cite{ma_adaptive_2021} (above), but Ko\c{c}anao\u{g}ullari et al. additionally include language models as a priori information for the MAB algorithm. 
In \cite{guo_calibration-free_2021}, the agent uses a variant of the UCB algorithm which interprets EEG signals as contextual information when choosing actions. 
Only two actions with a binary reward $r_{a_t}$ are available at each time step (the set of $K$ is two actions), respectively representing if the EEG signal had a P300 component or not.

\section{Discussion of Future use of Multi-Armed~Bandits in Brain-Computer~Interfaces}
\label{sec:BCI+MAB_future}
There are many possible uses for multi-armed bandits in BCI systems. 
Here, we present some directions for future research.

\subsection{Attention selection}
For users of hearing aids, it would be preferable, especially in a noisy environment, if the hearing aids would amplify only the attended sound source. 
Deciphering a particular sound source from a combination of sources is often referred to as the cocktail party problem. 
For the case of BCIs, there is research done on extracting the target source based on EEG measurements of the user's brain activity facilitating so called attention steering in hearing aids, \cite{alickovic_tutorial_2019}.

In a MAB formulation, each sound source corresponds to one action for the hearing aids. 
The reward for each action could be measured from the EEG data as Error potentials (ErrP) \cite{abiri_comprehensive_2019}, or the overall mental state \cite{nam_passive_2018}. 
The problem can be formulated in a few different ways based on different assumptions:

\begin{enumerate}[i) ]
	\item Within a limited time, the surrounding sound sources are the same, and the user keeps the same interest in them. Hence, the reward for each action is stationary, analogous to the original MAB formulation.
	\item The user can change their preferred sound source at any time, which can be modeled with non-stationary rewards, such as a switching bandit formulation. 
	One can assume as in \cite{hartland_change_2007} that it is only the best action that has a change in the reward, which means that the user can only lose interest in the target source, rather than hearing something else that gains their interest.
	\item Another approach would be to assume that sound sources can appear and disappear more or less randomly, which could be viewed as a mortal bandit problem as in \cite{chakrabarti_mortal_2008}.
\end{enumerate}

\subsection{Data for transfer learning}
A problem for BCI systems is the long calibration time due to the need for diverse data. 
Using data from previous sessions or persons and using transfer learning to adapt the old data to the current session is one solution \cite{lotte_signal_2015}. 
To find relevant data, one can among other approaches use tensor decomposition \cite{jeng_low-dimensional_2021}, Riemannian geometry \cite{khazem_minimizing_2021}, or a generic machine learning model \cite{jin_study_2020}. 
In \cite{gutierrez_multi-armed_2017} they use the classic MAB problem to find clusters of data in a big data set which increases the classification accuracy. 
The set of actions $K$, corresponds to the clusters, and the reward $r_{a_t}$ mirrors the classification accuracy when using training data from the selected cluster.
The ``one button BCI'' described in \cref{sec:one_button_BCI} aims at finding clusters of good data in an online calibration setting. Here, the clusters are for finding suitable data for transfer learning. 
The MAB problem formulation is similar in both cases despite the difference in application.

\subsection{Optimal calibration data}
Another solution than transfer learning to the problem with calibration time  \cite{lotte_signal_2015}  is to collect calibration data cleverly. 
Instead of collecting from all classes, as in uniform calibration, data could be collected from the class that would improve the classification accuracy the most.
Finding the optimal class could be formulated as a MAB problem where the set of actions $K$ represent the available classes, and the reward $r_{a_t}$ the gain in classification accuracy. 
Non-stationary rewards are a challenge in this setup since they will change with the current classification performance. 
Compared to the ``one button BCI'' described in \cref{sec:one_button_BCI}, the aim here is to have a ``multi-button BCI system'' using all classes for control, while the ``One button BCI system'' aims to find a single optimal class and solely use that one for control \cite{fruitet_bandit_2012,fruitet_automatic_2013}. 

\subsection{Best stopping time}
Another interesting aspect of the calibration phase raised in \cite{fruitet_bandit_2012} is to find the best stopping time. 
There are two variants of this formulation: 
\begin{enumerate}[i) ]
	\item The BCI system gets a limited time horizon $T$ of data samples to collect during the calibration and should reach a classification accuracy as high as possible by wisely collecting the data. 
	\item Assuming instead that there is a level of `good enough' classification accuracy  
	 the multi-armed bandit formulation can be used to minimize regret or decrease the time  $T$ for reaching the required accuracy level. 
\end{enumerate}

It is not evident that the multi-armed bandit formulation is the best way to solve either of these two problems, but it is worthwhile considering.

\section{Getting started with multi-armed bandits}
\label{sec:coding}

For most popular programming languages one can find examples of MABs \cite{github_multi-armed-bandit_nodate}. 
Among other ready to use packages are: ``SymPyBandits'' for Python \cite{SMPyBandits}, ``Bandits'' for Julia \cite{julia_bandits}, and ``Contextual'' for R \cite{van_emden_contextual_2020}. 
None of these packages are aimed at MABs in BCIs. 
Hence, we provide a brief Python example on BCI data that can act as a starting point for other researchers: \href{https://gitlab.control.lth.se/FridaH/mab_for_bci-public}{gitlab.control.lth.se/FridaH/mab\_for\_bci-public}.

\section{Conclusion}
\label{sec:conclusion}
Multi-armed bandits (MABs) have been used successfully in many fields, yet few applications for Brain-Computer Interfaces (BCIs) exist. 
Firstly, this review introduces MABs to the BCI field. 
Common algorithms to solve the classic MAB problem with stationary rewards include the $\epsilon$-greedy policy, the UCB algorithm, and Thompson sampling, all with the aim to balance the trade-off between exploration and exploitation of available actions. 
Secondly, the review highlights current research that interprets and solves BCI problems as MAB problems, prominently occurring in calibration optimization and real-time implementations of BCI systems. 
Finally, some suggestions are provided on further research directions in the intersection of MABs and BCIs.

\section*{Conflict of Interest Statement}

The authors declare that the research was conducted in the absence of any commercial or financial relationships that could be interpreted as a potential conflict of interest.

\section*{Author Contributions}
All authors have contributed to the conceptualization of the manuscript, manuscript revision, read, and approved the submitted version. FH wrote the first draft of the manuscript.

\section*{Funding}
This work was partially supported by the Wallenberg AI, Autonomous Systems and Software Program (WASP) funded by the Knut and Alice Wallenberg Foundation. All authors are also members of the ELLIIT Strategic Research Area.

\bibliographystyle{apalike}

\bibliography{main.bib}

\end{document}